\documentclass[journal]{IEEEtran}
\usepackage[T1]{fontenc}
\usepackage{cite}
\usepackage{url}
\usepackage{amsmath,amssymb,amsfonts}
\usepackage{graphicx}
\usepackage{textcomp}
\usepackage[table]{xcolor}
\usepackage{bm}
\usepackage{amsthm}
\usepackage{enumerate}
\usepackage{booktabs}
\usepackage{subfigure}
\usepackage{caption}
\usepackage{algorithmicx}
\usepackage[linesnumbered,ruled,vlined]{algorithm2e}
\usepackage{multirow}
\usepackage{array}
\definecolor{seagreen}{rgb}{0.18, 0.55, 0.34}
\definecolor{royalpurple}{rgb}{0.47,0.32,0.66}
\definecolor{brown(traditional)}{rgb}{0.59, 0.29, 0.0}
\definecolor{blue}{rgb}{0.3, 0.2, 0.9}
\usepackage[colorlinks,
            linkcolor=blue,
            anchorcolor=blue,
            citecolor=blue]{hyperref}

\DeclareMathOperator*{\argmin}{argmin}
\DeclareMathOperator*{\argmax}{argmax}

\begin{document}

\title{Frequency-Aware Continual Learning for Smart Contract Vulnerability Detection with Large Language Models}

\author{
Tenghui Huang, Jiawen Kang, \textit{Senior Member, IEEE}, Dongning Liu, \textit{Senior Member, IEEE}, \\Changyan Yi, \textit{Senior Member, IEEE}, Chengjun Cai, Anjia Yang, Li Li, and Dong In Kim,  \textit{Life Member, IEEE}

\thanks{
T. Huang and J. Kang are with the School of Automation, Guangdong University of Technology, Guangzhou 510006, China (e-mails: 3123000938@mail2.gdut.edu.cn, kavinkang@gdut.edu.cn).

D. Liu is with the School of Computer Science and Technology, Guangdong University of Technology, Guangzhou 510006, China (e-mail: liudn@gdut.edu.cn).

C. Yi is with the College of Computer Science and Technology, Nanjing University of Aeronautics and Astronautics, Nanjing 211106, China. (e-mail: changyan.yi@nuaa.edu.cn).

C. Cai is with the Department of Computer Science, City University of Hong Kong (Dongguan), Dongguan, Guangdong 518057, China (e-mail: chengjun.cai@cityudg.edu.cn).

Anjia Yang is with the College of Cyber Security, Jinan University,
Guangzhou, Guangdong 510632, China (e-mail: anjiayang@gmail.com).

L. Li is with the Guangdong Institute of Science and Technology Information, Guangzhou, Guangdong, China (e-mail: lichi12306@163.com).

Dong In Kim is with the Department of Electrical and Computer Engineering, Sungkyunkwan University, Suwon 16419, South Korea (e-mail: 
dongin@skku.edu).
}
\thanks{\textit{Corresponding author: Jiawen Kang.}}
}
\maketitle

\begin{abstract}
Smart contract vulnerability detection with Large Language Models (LLMs) faces three causally linked challenges. First, new vulnerability categories demand parameter-efficient adaptation, since full retraining is prohibitive for sequentially arriving tasks. Second, training per-task adapters on a shared backbone causes catastrophic forgetting of previously learned vulnerabilities. Third, the resulting multiplicity of adapters must be consolidated into a single model, since task identity is unknown at inference time. Each challenge arises directly from the solution to its predecessor, making an integrated framework essential. We propose a three-stage pipeline in which each stage addresses one challenge and feeds into the next. The adaptation stage uses Frequency-Aware Low-Rank Adaptation (FA-LoRA), which performs adaptation in the Fourier domain with per-frequency importance gates, requiring only 0.4\% trainable parameters while outperforming standard LoRA and QLoRA. The continual learning stage applies Forget-Aware Replay (FAR), which uses these frequency gates to estimate per-sample forgetting risk via loss dynamics and prioritizes vulnerable knowledge for rehearsal, achieving an average Micro-F1 of 0.8022 across sequential tasks. The deployment stage employs Anchor-Protected Progressive Merging (APPM), which exploits the asymmetric generalization produced by FAR training to identify the strongest-generalizing adapter as an anchor and consolidates all adapters into a single model via anchor-protected weighted merging with frequency-domain gate competition. APPM achieves a Micro-F1 of 0.8085, within 2.7\% of the independent per-task upper bound, at a merge cost of 156 ms and no additional runtime memory. Extensive experiments on the DIVE benchmark confirm that the proposed framework jointly addresses efficient adaptation, forgetting mitigation, and deployment consolidation for smart contract vulnerability detection in evolving blockchain ecosystems.

\end{abstract}


\section{Introduction}

\IEEEPARstart{S}{\MakeLowercase{mart}} contracts govern billions of dollars in decentralized digital assets on modern blockchains, functioning as the execution backbone of these ecosystems and making their security a matter of paramount importance. Vulnerabilities such as reentrancy, integer overflow, and access control flaws continue to cause severe financial losses despite extensive auditing efforts~\cite{kushwaha2022systematic}. Because deployed smart contracts are immutable by design, any vulnerability that escapes pre-deployment detection becomes permanently embedded and cannot be readily corrected afterward. Accurate vulnerability detection is therefore a foundational requirement for blockchain security~\cite{jiao2024survey}.

Large Language Models (LLMs) have shown strong capability in smart contract vulnerability detection by capturing complex program semantics and reasoning over security patterns~\cite{sun2024gptscan,liu2025propertygpt,chen2025chatgpt}. Existing LLM-based approaches, however, assume that all vulnerability categories are known at training time. In practice, smart contract platforms, protocols, and attack techniques continue to evolve, introducing new vulnerability classes over time~\cite{niyato2024aigx}. A model trained solely on historical data becomes increasingly incomplete as new attack patterns surface, creating a persistent gap between its detection capability and the actual threat landscape. A deployed detector must therefore incorporate newly discovered vulnerabilities while preserving previously acquired knowledge. Static models cannot handle emerging categories, full retraining is prohibitively expensive, and separate task-specific models incur excessive storage and inference overhead~\cite{yang2024model}.

Continual learning (CL) provides a natural solution by enabling a model to learn new tasks sequentially while retaining previously learned knowledge. Applying CL to LLM-based vulnerability detection, however, introduces three challenges that are causally linked. The first challenge is parameter-efficient adaptation. Full fine-tuning is prohibitively expensive for a stream of sequentially arriving tasks, yet per-task adapters offer a practical alternative by sharing a frozen backbone and requiring only a small fraction of parameters to be updated~\cite{hu2022lora,dettmers2023qlora}. The second challenge arises directly from this solution. Sequentially training per-task adapters on a shared backbone inevitably overwrites parameters critical to previously learned vulnerabilities, causing catastrophic forgetting~\cite{kirkpatrick2017ewc,buzzega2020dark}. The third challenge follows from the remedy to forgetting. Mitigating forgetting via CL mechanisms produces multiple task-specific adapters, but maintaining separate adapters requires knowing the task identity at inference time, which is impractical in deployment, while naively merging them causes destructive parameter interference. Each challenge is a direct consequence of solving the one before it, and no single challenge can be resolved in isolation.

To address this causal chain, we propose a three-stage framework that tackles each challenge in sequence. The adaptation stage employs Frequency-Aware Low-Rank Adaptation (FA-LoRA), which learns compact adapters in the Fourier frequency domain with per-frequency importance gates. FA-LoRA requires only 0.4\% trainable parameters while outperforming standard LoRA and QLoRA. The continual learning stage uses Forget-Aware Replay (FAR), which applies the frequency gates from FA-LoRA to estimate per-sample forgetting risk via loss dynamics and prioritizes high-risk samples for rehearsal, achieving an average Micro-F1 of 0.8022 across sequential tasks. The deployment stage applies Anchor-Protected Progressive Merging (APPM), which exploits the asymmetric generalization that FAR training produces to identify the strongest adapter as an anchor and consolidate all adapters into a single model via anchor-protected weighted merging and frequency-domain gate competition, achieving a Micro-F1 of 0.8085 within 2.7\% of the independent per-task upper bound at a merge cost of only 156 milliseconds.

The three stages share a design dependency. FAR relies on the frequency gates produced by FA-LoRA to estimate which knowledge dimensions are most vulnerable to forgetting. APPM depends on two upstream outputs. The gate parameters from FA-LoRA enable frequency-domain competition during merging, while the asymmetric training outcomes from FAR provide the basis for anchor selection. Since later adapters trained via FAR accumulate knowledge from all previous tasks, they naturally generalize more strongly and serve as stronger anchors. This creates a reinforcing cycle in which better CL training produces better anchors, and better anchors yield better merged models.

The main contributions of this paper are summarized as follows.

\begin{itemize}

\item We propose a unified continual learning framework for LLM-based smart contract vulnerability detection that integrates efficient adaptation, forgetting mitigation, and deployment consolidation into a cohesive three-stage pipeline. The framework jointly addresses parameter efficiency, catastrophic forgetting, and multi-task inference, providing an end-to-end solution from sequential task learning to unified model deployment.

\item We propose FA-LoRA, a frequency-aware parameter-efficient fine-tuning method that performs low-rank adaptation in the Fourier domain. FA-LoRA requires only 0.4\% trainable parameters while consistently outperforming standard LoRA and QLoRA.

\item We develop FAR, a continual learning algorithm that prioritizes replay according to per-sample forgetting risk estimated from loss dynamics, achieving the best continual learning performance among compared methods on sequential vulnerability detection tasks.

\item We propose APPM, a training-data-free adapter merging strategy that consolidates multiple task-specific adapters into a single deployment model. APPM achieves performance within 2.7\% of the independent per-task upper bound with millisecond-level merging time and no additional runtime memory overhead.

\end{itemize}

The remainder of this paper is organized as follows. Section~\ref{sec:related} reviews related work. Section~\ref{sec:system} presents the proposed framework and its architecture. Section~\ref{sec:falora} introduces FA-LoRA. Section~\ref{sec:cl} describes the continual learning framework and FAR. Section~\ref{sec:appm} presents APPM. Section~\ref{sec:experiments} reports experimental results, and Section~\ref{sec:conclusion} concludes the paper.

\section{Related Work}
\label{sec:related}

\subsection{Smart Contract Vulnerability Detection}

Smart contract vulnerability detection has traditionally relied on static analysis and dynamic testing. Static analysis tools such as Slither~\cite{feist2019slither}, Securify~\cite{tsankov2018securify}, and SmartCheck~\cite{tikhomirov2018smartcheck} detect vulnerabilities through handcrafted rules, data-flow analysis, and formal compliance checking. Fuzzing-based approaches, including ContractFuzzer~\cite{jiang2018contractfuzzer} and Smartian~\cite{choi2021smartian}, explore execution paths by generating transaction inputs guided by contract interfaces and program dependencies. More recently, SmartAxe~\cite{liao2024smartaxe} targets cross-chain bridge vulnerabilities through fine-grained static analysis, and Satellite~\cite{zheng2025satellite} detects vulnerabilities caused by subcontract misuse. Although effective for predefined vulnerability patterns, these methods generally suffer from limited semantic understanding and poor generalization to previously unseen attack types~\cite{jiao2024survey}.

Recent studies have demonstrated the effectiveness of LLMs for smart contract security analysis. GPTScan~\cite{sun2024gptscan} combines GPT-4 with static analysis to improve vulnerability confirmation, while PropertyGPT~\cite{liu2025propertygpt} incorporates property-guided reasoning to enhance detection coverage. Other studies fine-tune open-source LLMs for vulnerability classification~\cite{boi2024smart,ma2025combining,yu2025smartllama,huang2026paravul}, and recent work further leverages LLMs to augment smart contract decompiler output through semantic recovery~\cite{zheng2025llmdecompiler}. However, these methods assume a static vulnerability taxonomy and lack the ability to adapt to newly emerging vulnerability categories.

\subsection{Parameter-Efficient Fine-Tuning for LLMs}

PEFT has become the standard approach for efficient LLM adaptation, with training data transparency emerging as a growing concern~\cite{yang2026detecting}. LoRA~\cite{hu2022lora} freezes the backbone and injects trainable low-rank matrices, drastically cutting the parameter count. QLoRA~\cite{dettmers2023qlora} further combines low-rank adaptation with 4-bit quantization for commodity-hardware fine-tuning. More recent variants incorporate sparse adapters~\cite{ding2023sparse,bhardwaj2024sparse}, adaptive rank allocation, and learnable gating.

More recently, frequency-domain PEFT methods have attracted increasing attention. FourierFT~\cite{gao2024fourierft} learns parameter updates in the Fourier domain, while FouRA~\cite{borse2024foura} further demonstrates the advantages of frequency-domain representations for multi-task adapter merging. However, existing methods typically employ fixed frequency selection strategies and focus primarily on single-task adaptation. Their applicability to continual learning scenarios remains largely unexplored.

\subsection{Continual Learning for Sequential Tasks}

CL aims to enable models to acquire new knowledge without forgetting previously learned tasks~\cite{wang2024comprehensive,shi2024continual}. Existing approaches can generally be categorized into replay-based, regularization-based, and architectural methods. Replay-based methods, such as DER~\cite{buzzega2020dark}, mitigate forgetting by revisiting samples from previous tasks. Regularization methods, including EWC~\cite{kirkpatrick2017ewc}, Online EWC~\cite{schwarz2018progress}, and MAS~\cite{aljundi2018mas}, preserve important parameters by constraining weight updates. Architectural approaches allocate dedicated model capacity for different tasks to reduce interference~\cite{lopez2017gem}.

Although continual learning has achieved considerable success in computer vision and natural language processing, its application to smart contract vulnerability detection remains largely unexplored. Related work on federated unlearning~\cite{kang2024federatedunlearning} and one-shot federated learning~\cite{kang2024oneshot} addresses knowledge retention in distributed settings, though the sequential multi-label scenario considered in this paper poses distinct challenges. Unlike conventional class-incremental benchmarks, vulnerability detection is inherently multi-label, and vulnerability categories often exhibit strong semantic correlations, making catastrophic forgetting more challenging to address.

\subsection{Model Merging for Multi-Task LLMs}

Model merging aims to consolidate multiple task-specific models into a single model without additional training, thereby reducing deployment cost~\cite{yang2024model}. Representative weight-space approaches include Task Arithmetic~\cite{ilharco2023task}, TIES-Merging~\cite{yadav2023ties}, DARE~\cite{yu2024dare}, and RegMean~\cite{jin2023regmean}. Recent studies further investigate Fisher-weighted merging~\cite{matena2022merging}, frequency-domain merging~\cite{borse2024foura,zhao2025cmlora}, and SVD-based LoRA composition~\cite{chen2025lora}.

Most existing merging methods assume that all task adapters contribute equally during merging and therefore ignore the asymmetric knowledge accumulation introduced by continual learning. They are not designed for deployment scenarios where later tasks inherit knowledge acquired from earlier ones.

Existing studies have advanced smart contract vulnerability detection, parameter-efficient fine-tuning, continual learning, and model merging as separate threads. Smart contract detectors assume static vulnerability distributions, PEFT methods prioritize efficient adaptation without supporting continual learning, CL methods focus on mitigating forgetting without considering deployment efficiency, and model merging approaches overlook the sequential knowledge accumulation inherent in continual learning. An integrated framework that jointly addresses adaptation efficiency, forgetting mitigation, and deployment consolidation is still lacking.

\section{Proposed Framework}
\label{sec:system}


\begin{figure}[]
\centering
\includegraphics[width=0.98\linewidth]{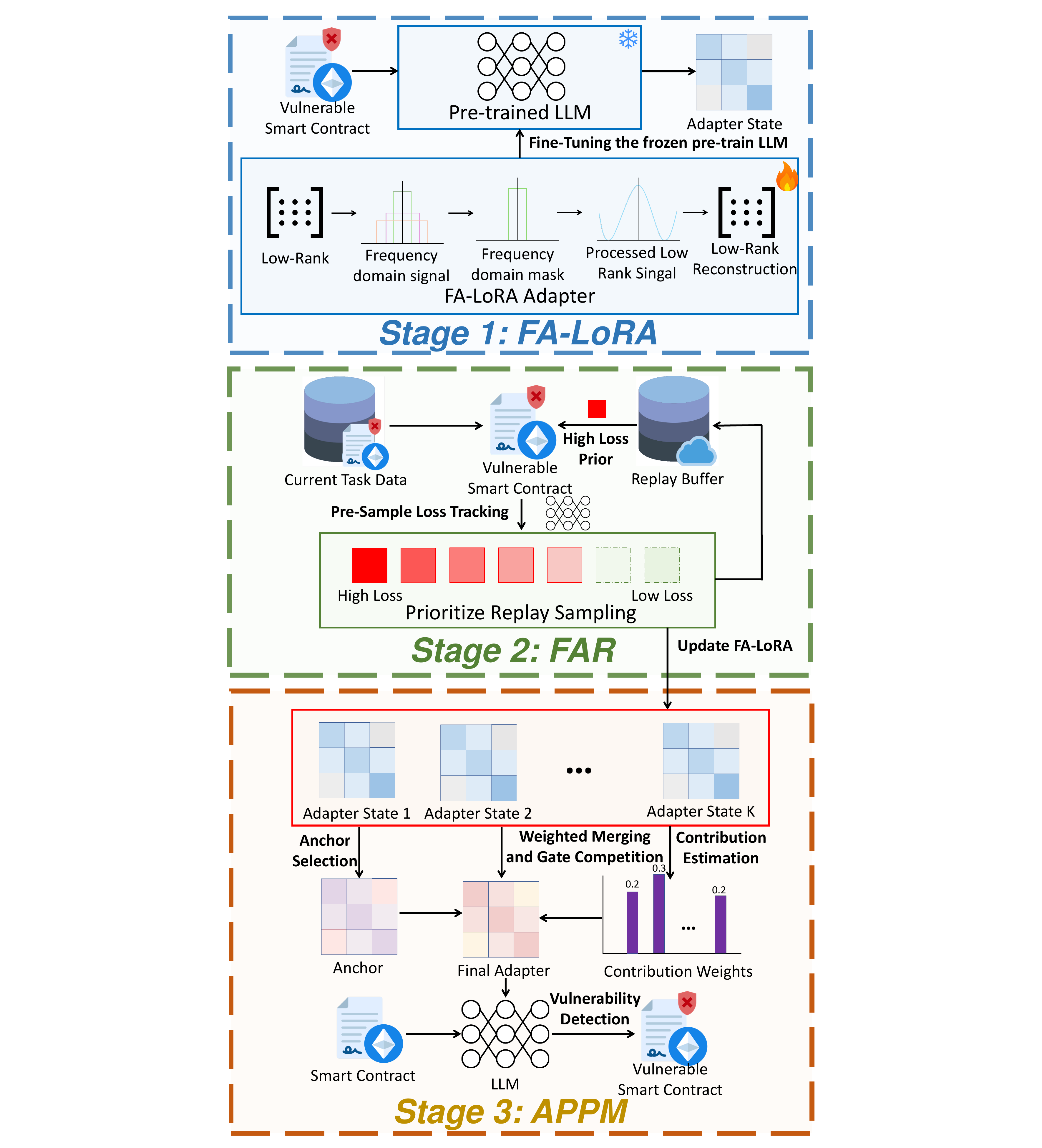}
\caption{Overview of the proposed continual learning framework for LLM-based smart contract vulnerability detection. }\label{fig:framework}
\end{figure}

\subsection{Problem Setting and Design Rationale}


In a production deployment scenario, a smart contract vulnerability detector faces a stream of temporally ordered vulnerability tasks $\mathcal{T}_1, \mathcal{T}_2, \ldots, \mathcal{T}_K$ with evolving vulnerability patterns. The detector must adapt to each new task with minimal computational overhead, since full retraining of LLMs is prohibitively expensive. At the same time, accuracy on previously learned vulnerability categories must be preserved against catastrophic forgetting, and the deployed model must remain a single set of parameters detecting all learned vulnerabilities without task-identity information at inference time. These three requirements pull in opposing directions: mitigating forgetting typically requires storing historical data or parameters, conflicting with efficient adaptation, while maintaining task-specific models violates unified deployment. Our framework resolves this tension through three integrated components, as illustrated in Fig.~\ref{fig:framework}.

\subsection{Three-Stage Pipeline Architecture}

The framework operates as a three-stage pipeline.

\textbf{Stage 1: Efficient Adaptation via FA-LoRA.} When a new vulnerability task $\mathcal{T}_k$ arrives, the system fine-tunes only a lightweight FA-LoRA adapter while keeping the pre-trained LLM backbone frozen. The adapter introduces frequency-domain gating and sparse frequency selection, requiring only 0.4\% trainable parameters, with each task producing a compact adapter state of negligible storage cost.

\textbf{Stage 2: Forgetting Mitigation via FAR.} During sequential training across tasks, the system maintains a fixed-capacity replay buffer. After each task completes, per-sample forgetting risk is estimated from recent loss dynamics. During subsequent training, samples with higher forgetting risk receive proportionally higher replay probability, ensuring that vulnerable knowledge receives prioritized rehearsal. This loss-dynamics-driven prioritization achieves stronger retention than uniform replay at identical buffer capacity.

\textbf{Stage 3: Unified Deployment via APPM.} After all $K$ tasks have been learned, the $K$ task-specific FA-LoRA adapters are consolidated into a single unified adapter through APPM. The merging process operates entirely in parameter space without accessing training data. It first identifies the best-generalizing adapter as the anchor, then performs norm-weighted parameter aggregation for LoRA matrices and anchor-boosted frequency-domain competition for gate parameters. The resulting merged adapter supports multi-task inference with accuracy approaching the independent per-task upper bound, while introducing no runtime overhead.

\subsection{Component Interaction}

The three components are designed so that each stage produces outputs used by downstream stages. The frequency-domain parameterization of FA-LoRA provides adapters whose spectral structure directly supports two downstream operations. First, the frequency gate parameters indicate which knowledge dimensions are most vulnerable to forgetting, informing the loss-dynamics-driven prioritization in FAR. Second, the same gate parameters enable the frequency-domain competition mechanism of APPM, where each frequency bin is allocated to the task that activates it most strongly. Conventional LoRA adapters lack explicit frequency modeling and therefore cannot support either operation.

In turn, the anchor protection mechanism of APPM exploits the sequential knowledge accumulation inherent to continual learning, where later adapters trained on richer knowledge serve as stronger anchors for the merged representation. This creates a reinforcing cycle in which better CL algorithms produce better anchors, and better anchors yield better merged models. The pipeline maintains a single frozen architecture throughout, stores only compact adapter states, and requires only milliseconds of CPU time for final merging.

\section{Frequency-Aware Low-Rank Adaptation}
\label{sec:falora}


\subsection{Design Motivation}

Low-Rank Adaptation (LoRA)~\cite{hu2022lora} has emerged as an effective PEFT technique by introducing trainable low-rank updates into frozen large language models. Specifically, given a pre-trained weight matrix $\boldsymbol{W}\in\mathbb{R}^{d\times d}$, LoRA represents its adaptation as:

\begin{equation}
    \Delta\boldsymbol{W}
    =
    \boldsymbol{U}\boldsymbol{V},
\end{equation}

where $\boldsymbol{U}\in\mathbb{R}^{d\times r}$ and
$\boldsymbol{V}\in\mathbb{R}^{r\times d}$ with $r\ll d$. The forward
propagation is formulated as:

\begin{equation}
    \boldsymbol{h}
    =
    \boldsymbol{x}\boldsymbol{W}
    +
    \alpha\boldsymbol{x}(\boldsymbol{U}\boldsymbol{V}),
\end{equation}

where $\alpha$ is a scaling factor and $\boldsymbol{W}$ remains frozen.

Although LoRA significantly reduces trainable parameters, its low-rank update
mechanism treats all adaptation components equally. This uniform optimization
may limit the capability of LoRA in continual learning scenarios, where
different tasks may require different adaptation patterns. In particular,
sequential vulnerability detection tasks may contain both general security
knowledge and task-specific vulnerability characteristics, making it important
to selectively preserve informative adaptation signals.

Recent studies on frequency-domain parameter-efficient adaptation
~\cite{gao2024fourierft,borse2024foura} suggest that neural representations
contain diverse frequency components with different information characteristics.
Motivated by this observation, FA-LoRA introduces frequency-domain
decomposition into LoRA updates and adaptively controls frequency components
through learnable gating and sparse frequency selection.

\subsection{Frequency-Aware Adapter Architecture}

\begin{figure}[t]
    \centering
    \includegraphics[width=0.98\linewidth]{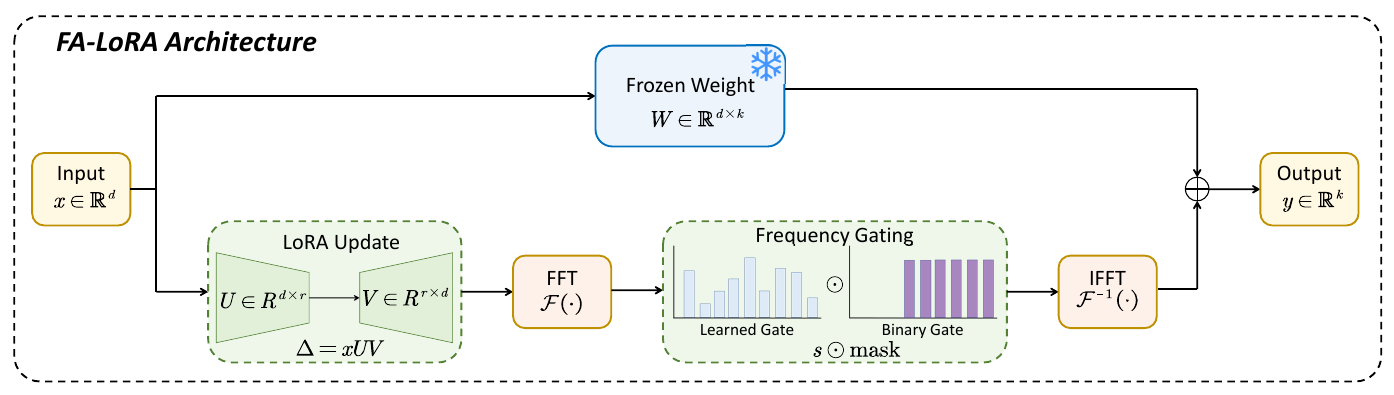}
    \caption{Architecture of FA-LoRA. }
    \label{fig:falora}
\end{figure}


The overall architecture of FA-LoRA is illustrated in Fig.~\ref{fig:falora}. Given an input representation $\boldsymbol{x}\in\mathbb{R}^{n\times d}$, FA-LoRA first generates the standard low-rank adaptation signal:

\begin{equation}
    \Delta\boldsymbol{H}
    =
    \boldsymbol{x}(\boldsymbol{U}\boldsymbol{V}),
\end{equation}

where $\Delta\boldsymbol{H}$ denotes the LoRA-induced representation update.
Instead of directly applying the low-rank update in the original feature
space, FA-LoRA transforms it into the frequency domain using the discrete
DFT:

\begin{equation}
    \boldsymbol{F}_{\Delta H}
    =
    \mathcal{F}(\Delta\boldsymbol{H}),
\end{equation}

where $\mathcal{F}(\cdot)$ represents the Fourier transform operation.
The frequency representation $\boldsymbol{F}_{\Delta H}$ explicitly separates
the adaptation signal into different frequency components.

To adaptively adjust the contribution of different frequency components, FA-LoRA
introduces a learnable frequency gate:

\begin{equation}
    \boldsymbol{s}
    =
    \sigma(\boldsymbol{g}),
\end{equation}

where $\boldsymbol{g}$ denotes trainable frequency importance parameters and
$\sigma(\cdot)$ is the sigmoid activation function. The obtained gate vector
$\boldsymbol{s}\in[0,1]^r$ provides soft frequency-wise modulation.

Furthermore, FA-LoRA applies sparse frequency selection to focus adapter capacity on the most informative components. A binary frequency mask is defined as:

\begin{equation}
\mathrm{mask}_i
=
\begin{cases}
1, & i\in\mathcal{K},\\
0, & otherwise,
\end{cases}
\end{equation}

where $\mathcal{K}$ represents the selected frequency indices:

\begin{equation}
    \mathcal{K}
    =
    \mathrm{TopK}
    (|\boldsymbol{s}|,\lceil\gamma r\rceil).
\end{equation}

Here, $\gamma$ controls the retained frequency ratio. Following previous observations that high-frequency components often contain fine-grained adaptation signals beneficial for task-specific knowledge adjustment, we retain only the most informative high-frequency components.

The gated frequency representation is calculated as:

\begin{equation}
    \boldsymbol{F}_{FA}
    =
    \boldsymbol{F}_{\Delta H}
    \odot
    (\boldsymbol{s}\odot\mathrm{mask}),
\end{equation}

where $\odot$ is the element-wise product. The final FA-LoRA update is reconstructed via inverse FFT:
\begin{equation}
    \Delta\boldsymbol{H}_{FA}
    =
    \mathcal{F}^{-1}
    (\boldsymbol{F}_{FA}),
\end{equation}

where $\mathcal{F}^{-1}(\cdot)$ denotes the inverse Fourier transform.
Therefore, the complete forward computation of a FA-LoRA-enhanced transformer
layer is:

\begin{equation}
    \boldsymbol{h}
    =
    \boldsymbol{x}\boldsymbol{W}
    +
    \alpha\Delta\boldsymbol{H}_{FA}.
\end{equation}

FA-LoRA is applied to the query and value projection matrices $\boldsymbol{W}_q$ and $\boldsymbol{W}_v$ of each transformer layer, following the conventional LoRA configuration. Algorithm~\ref{alg:falora} summarizes the forward propagation procedure of FA-LoRA.

\begin{algorithm}[t]
\caption{FA-LoRA Forward Pass (Single Layer)}
\label{alg:falora}
\DontPrintSemicolon
\SetAlgoLined

Initialize frozen weight $\boldsymbol{W} \in \mathbb{R}^{d \times d}$.

Initialize low-rank parameters $\boldsymbol{U} \in \mathbb{R}^{d \times r}, \boldsymbol{V} \in \mathbb{R}^{r \times d}$.

Initialize frequency gate $\boldsymbol{g} \in \mathbb{R}^r$ and retention ratio $\gamma$.

\For{\rm{each input} $\boldsymbol{x} \in \mathbb{R}^{n \times d}$}
{
    \textcolor{blue}{\textit{\#\#\# Frozen Projection \#\#\#}}

    Compute the base output: $\boldsymbol{z} \leftarrow \boldsymbol{x}\boldsymbol{W}$.

    \textcolor{blue}{\textit{\#\#\# Low-Rank Adapter \#\#\#}}

    Compute the low-rank update: $\Delta\boldsymbol{H} \leftarrow \boldsymbol{x}(\boldsymbol{U}\boldsymbol{V})$.

    \textcolor{blue}{\textit{\#\#\# Frequency-Domain Gating \#\#\#}}

    Transform to frequency domain: $\boldsymbol{F}_{\Delta H} \leftarrow \mathcal{F}(\Delta\boldsymbol{H})$.

    Compute gate activation: $\boldsymbol{s} \leftarrow \sigma(\boldsymbol{g})$.

    Select top-$k$ frequencies: $\mathcal{K} \leftarrow \mathrm{TopK}(|\boldsymbol{s}|, \lceil\gamma r\rceil)$.

    Construct binary mask: $\mathrm{mask} \leftarrow \mathbb{1}[i \in \mathcal{K}]$.

    Apply gated sparse mask: $\boldsymbol{F}_{FA} \leftarrow \boldsymbol{F}_{\Delta H} \odot (\boldsymbol{s} \odot \mathrm{mask})$.

    \textcolor{blue}{\textit{\#\#\# Inverse Transform \#\#\#}}

    Reconstruct adapter output: $\Delta\boldsymbol{H}_{FA} \leftarrow \mathcal{F}^{-1}(\boldsymbol{F}_{FA})$.

    \textcolor{blue}{\textit{\#\#\# Final Output \#\#\#}}

    Combine outputs: $\boldsymbol{h} \leftarrow \boldsymbol{z} + \alpha \Delta\boldsymbol{H}_{FA}$.
}

\end{algorithm}

\subsection{Parameter Efficiency}

FA-LoRA introduces only the low-rank matrices and frequency gate parameters
while keeping the original LLM parameters frozen. For each adapted linear
layer, the number of trainable parameters is:

\begin{equation}
    N_{\mathrm{FA-LoRA}}
    =
    2dr+r,
\end{equation}

where $r$ is the LoRA rank and $d$ is the hidden dimension.

Compared with full fine-tuning requiring $d^2$ trainable parameters for each linear layer, FA-LoRA reduces the optimization space from quadratic to linear complexity with respect to the hidden dimension. requiring only a small fraction of the total model parameters to be trainable.

This substantial parameter reduction enables efficient continual fine-tuning
while preserving the expressive capability of large language models.

\subsection{Training Objective}

For multi-label smart contract vulnerability detection with $L$ vulnerability
categories, FA-LoRA is optimized using the binary cross-entropy objective:

\begin{equation}
\mathcal{L}
=
-\frac{1}{L}
\sum_{i=1}^{L}
\left[
y_i\log\hat{y}_i
+
(1-y_i)\log(1-\hat{y}_i)
\right],
\end{equation}

where $y_i\in\{0,1\}$ is the ground-truth label and $\hat{y}_i$ is the predicted probability for vulnerability category $i$. During training, only $\boldsymbol{U}$, $\boldsymbol{V}$, and $\boldsymbol{g}$ are updated while all pre-trained LLM parameters remain frozen.

Before evaluating FA-LoRA in continual learning scenarios, we first validate
its effectiveness as an independent PEFT method. Section~\ref{sec:experiments}
compares FA-LoRA with representative PEFT baselines, including LoRA~\cite{hu2022lora}, QLoRA~\cite{dettmers2023qlora},
SLoRA~\cite{huang2026paravul}, FourierFT~\cite{gao2024fourierft}, and WaRA~\cite{heidari2025wara}, on the complete DIVE dataset.

\section{Continual Learning for Sequential Vulnerability Tasks}
\label{sec:cl}


\subsection{Problem Formulation}
\label{sec:cl-problem}

In real-world auditing, vulnerability knowledge evolves over time. A vulnerability category can only be labeled after its public discovery, so a deployed model should only access patterns known before that time point. This temporal constraint leads directly to a continual learning formulation.

Given a chronologically ordered contract corpus, we sort all contracts according to their deployment or commit timestamps and divide them into $K$ consecutive time intervals:

\begin{equation}
\mathcal{I}_k=[t_{k-1},t_k), \quad k=1,\ldots,K .
\end{equation}

Each interval defines a continual learning task:

\begin{equation}
\mathcal{T}_k=
\mathcal{D}_k
=
\{
(\boldsymbol{x}_i,\boldsymbol{y}_i)
\}_{i=1}^{N_k},
\end{equation}

where $\boldsymbol{x}_i$ denotes a smart contract representation and
$\boldsymbol{y}_i\in\{0,1\}^{L}$ is the multi-label vulnerability annotation
over $L$ vulnerability categories.

Under the temporal partition, early tasks contain vulnerability patterns known at earlier stages, while later tasks introduce newly emerging vulnerability characteristics. The concrete benchmark construction adopted in our evaluation is detailed in Section~\ref{sec:experiments}.

The objective of continual learning is to sequentially optimize the model over
tasks:

\begin{equation}
\mathcal{T}_1
\rightarrow
\mathcal{T}_2
\rightarrow
\cdots
\rightarrow
\mathcal{T}_K ,
\end{equation}

while preserving performance on previously learned tasks. Let $\mathrm{F1}_{i,j}$ denote the performance on task $j$ after learning task $i$, and let FWT and BWT denote Forward Transfer and Backward Transfer, respectively:

\begin{equation}
\mathrm{FWT}
=
\frac{1}{K-1}
\sum_{k=2}^{K}
(
\mathrm{F1}_{k-1,k}
-
\mathrm{F1}_{base,k}
),
\end{equation}

\begin{equation}
\mathrm{BWT}
=
\frac{1}{K-1}
\sum_{k=2}^{K}
\frac{1}{k-1}
\sum_{j=1}^{k-1}
(
\mathrm{F1}_{k,j}
-
\mathrm{F1}_{j,j}
).
\end{equation}

Positive FWT indicates beneficial knowledge transfer to future tasks, whereas
negative BWT reflects performance degradation caused by catastrophic
forgetting.

\subsection{Estimating Forgetting from Loss Dynamics}
\label{sec:cl-forget}

While replay-based methods alleviate forgetting, they treat all stored samples uniformly. Yet forgetting is highly sample-dependent. Some samples stay robust across tasks, while others degrade sharply. Uniform replay thus wastes limited buffer capacity on knowledge already retained.


To this end, FAR estimates per-sample forgetting risk from recent training losses. Each buffer entry stores the input $\boldsymbol{x}_i$, the vulnerability label $\boldsymbol{y}_i$, and a forgetting indicator $\ell_i$ given by the sample-wise binary cross-entropy loss:

\begin{equation}
\ell_i
=
-\frac{1}{L}
\sum_{m=1}^{L}
\left[
y_{i,m}\log(\hat{y}_{i,m})
+
(1-y_{i,m})
\log(1-\hat{y}_{i,m})
\right],
\label{eq:sample_loss}
\end{equation}

where $y_{i,m}$ is the ground-truth label for category $m$ and $\hat{y}_{i,m}$ is the current model prediction. After each task $\mathcal{T}_k$, all buffered samples are re-evaluated through an inference pass and their loss values updated via Eq.~\eqref{eq:sample_loss}. These losses are used only to estimate forgetting risk and assign replay probabilities, without adding any optimization objective.

\subsection{Forget-Aware Replay Sampling}
\label{sec:cl-sampling}

During replay, samples are selected according to a temperature-scaled
probability distribution:

\begin{equation}
p_i
=
\frac{
\exp(\ell_i/\tau)
}
{
\sum_{j=1}^{|\mathcal{M}|}
\exp(\ell_j/\tau)
},
\label{eq:forgetreplay}
\end{equation}

where $\tau$ controls the prioritization intensity. A smaller $\tau$ concentrates replay probability on high-loss samples, while a larger $\tau$ flattens the distribution toward uniform sampling.

The proposed strategy has two advantages. First, it is task-agnostic because
the sampling criterion depends only on prediction difficulty rather than task
identity. Second, it is self-adaptive because successfully rehearsed samples
receive lower future probabilities.

\begin{figure}[t]
    \centering
    \includegraphics[width=0.98\linewidth]{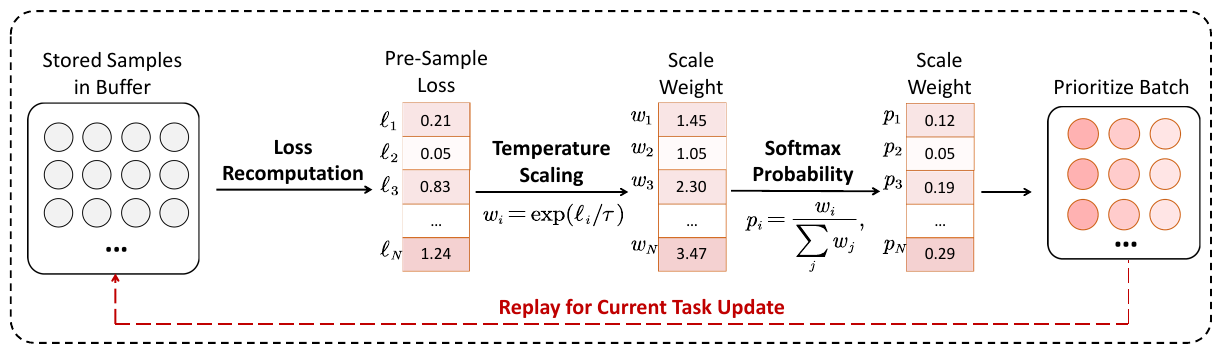}
    \caption{The FAR mechanism performing forget-aware prioritization via per-sample loss recomputation and temperature-scaled softmax sampling.}
    \label{fig:forget_replay}
\end{figure}

Algorithm~\ref{alg:forgetreplay} summarizes the sampling procedure. Fig.~\ref{fig:forget_replay} illustrates the overall workflow of FAR.

\begin{algorithm}[t]
\caption{Forget-Aware Replay Sampling}
\label{alg:forgetreplay}
\DontPrintSemicolon
\SetAlgoLined

Initialize replay buffer $\mathcal{M} = \{(\boldsymbol{x}_i, \boldsymbol{y}_i, \ell_i)\}_{i=1}^{|\mathcal{M}|}$.

Set temperature $\tau$ and replay batch size $B_r$.

    \textcolor{blue}{\textit{\#\#\# Compute Selection Weights \#\#\#}}

    \For{each sample $i$ in $\mathcal{M}$}
    {
        Compute loss-based weight: $w_i \leftarrow \exp(\ell_i / \tau)$.
    }

    \textcolor{blue}{\textit{\#\#\# Prioritized Sampling \#\#\#}}

    Normalize to softmax probabilities: $p_i \leftarrow w_i / \sum_j w_j$.

    Draw $B_r$ samples from $\mathcal{M}$ with probabilities $\{p_i\}$ to form replay batch $\mathcal{B}_r$.

\Return{$\mathcal{B}_r$}

\end{algorithm}

\subsection{Overall Training Objective}
\label{sec:cl-objective}

The complete FAR procedure can be formalized as follows. When task $\mathcal{T}_k$ arrives, the trainable FA-LoRA parameters $\boldsymbol{\Theta}$ are optimized against a combined objective that merges the current-task loss with a rehearsal loss over the prioritized replay batch:

\begin{equation}
\begin{split}
\mathcal{L}_{\mathrm{CL}}^{(k)}(\boldsymbol{\Theta})
=
&
\frac{1}{|\mathcal{T}_k|}
\sum_{(\boldsymbol{x}_i,\boldsymbol{y}_i)\in\mathcal{T}_k}
\ell(\boldsymbol{x}_i,\boldsymbol{y}_i;\boldsymbol{\Theta})
\\
&+
\lambda
\frac{1}{|\mathcal{B}_r|}
\sum_{(\boldsymbol{x}_j,\boldsymbol{y}_j)\in\mathcal{B}_r}
\ell(\boldsymbol{x}_j,\boldsymbol{y}_j;\boldsymbol{\Theta}),
\end{split}
\label{eq:cl_objective}
\end{equation}

where $\mathcal{B}_r\subseteq\mathcal{M}$ is the replay batch drawn with the probabilities $\{p_i\}$ of Eq.~\eqref{eq:forgetreplay}, and $\lambda$ balances adaptation to the new task against retention of previously learned knowledge, with its value determined by the replay batch ratio in Table~\ref{tab:training_config}.

The forgetting risk that drives the prioritization in Eq.~\eqref{eq:forgetreplay} is quantified by the change of the prediction loss of each buffered sample across consecutive task boundaries. After task $\mathcal{T}_k$ is learned, the forgetting risk of sample $i$ is defined as

\begin{equation}
\Delta\ell_i^{(k)}
=
\ell_i^{(k)}
-
\ell_i^{(k-1)},
\label{eq:forget_risk}
\end{equation}

where $\ell_i^{(k)}$ is the per-sample loss of Eq.~\eqref{eq:sample_loss} recomputed after training on $\mathcal{T}_k$, and the loss recorded upon the entry of the sample into the buffer serves as its baseline. A large positive $\Delta\ell_i^{(k)}$ indicates that the knowledge carried by sample $i$ is being actively overwritten, so the sample receives a higher replay probability in subsequent tasks.

Finally, the temperature-scaled distribution of Eq.~\eqref{eq:forgetreplay} admits a principled interpretation: it is the closed-form solution of an entropy-regularized maximization of the expected rehearsal loss,

\begin{equation}
\{p_i\}
=
\arg\max_{\{p_i\}}
\left[
\sum_{i=1}^{|\mathcal{M}|}
p_i\,\ell_i
-
\tau
\sum_{i=1}^{|\mathcal{M}|}
p_i \log p_i
\right],
\label{eq:entropy_replay}
\end{equation}


subject to $\sum_{i=1}^{|\mathcal{M}|} p_i = 1$. This concentrates rehearsal on high-risk samples, while the entropy term controlled by $\tau$ prevents the distribution from collapsing onto a single sample. Taken together, Eqs.~\eqref{eq:cl_objective}--\eqref{eq:entropy_replay} fully specify FAR, in which forgetting risk is estimated from loss dynamics, converted into sampling probabilities, and injected into the training objective through prioritized rehearsal.

\section{Anchor-Protected Progressive Merging for Unified Inference}
\label{sec:appm}



\subsection{Problem Definition}

Assume that continual learning produces $K$ task-specific FA-LoRA adapters

\begin{equation}
\mathcal{A}
=
\{
\boldsymbol{\Theta}^{(1)},
\boldsymbol{\Theta}^{(2)},
\ldots,
\boldsymbol{\Theta}^{(K)}
\},
\end{equation}

where

\begin{equation}
\boldsymbol{\Theta}^{(k)}
=
\{
(\boldsymbol{U}^{(k)}_l,
\boldsymbol{V}^{(k)}_l,
\boldsymbol{g}^{(k)}_l)
\}_{l=1}^{L}
\end{equation}

denotes the FA-LoRA parameters of task $k$ spanning $L$ transformer layers. APPM produces a single unified adapter

\begin{equation}
\boldsymbol{\Theta}^{*}
=
\Phi
(
\boldsymbol{\Theta}^{(1)},
\ldots,
\boldsymbol{\Theta}^{(K)}
),
\end{equation}

where $\Phi(\cdot)$ denotes the merging operator. The merging process is
performed entirely in parameter space without accessing any training samples,
making APPM a post-training data-free adapter merging method.

\begin{figure}[t]
    \centering
    \includegraphics[width=0.98\linewidth]{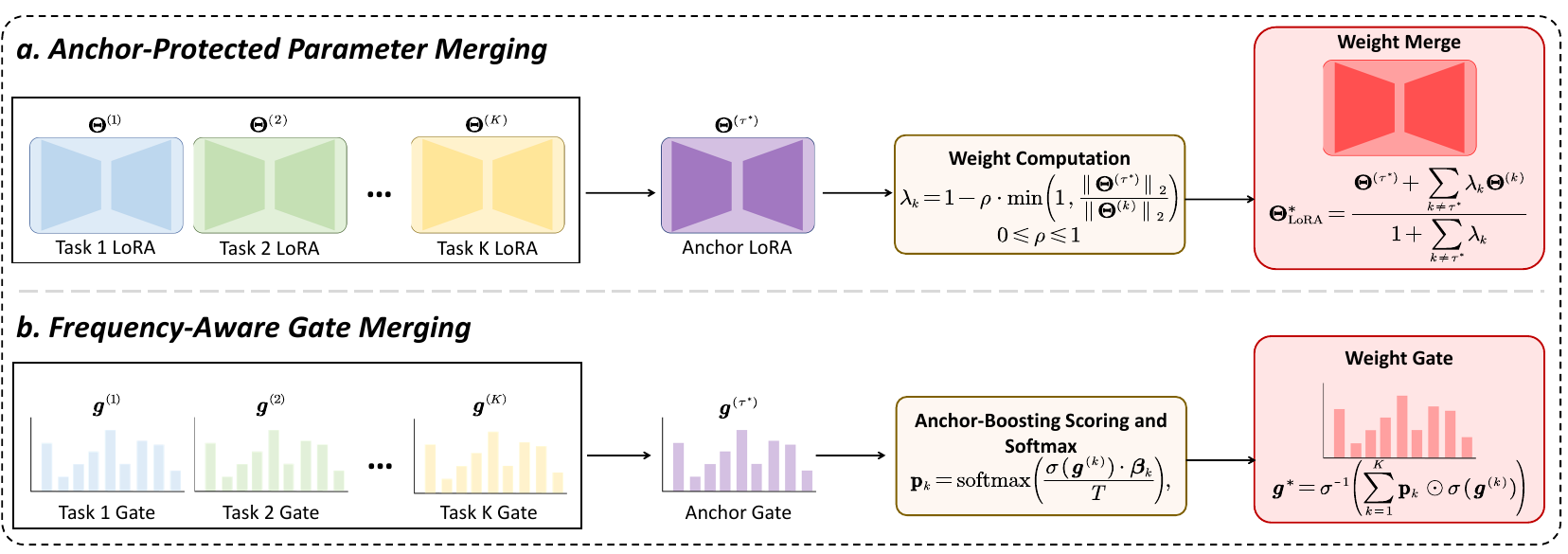}
    \caption{The two-stage APPM workflow.}
    \label{fig:appm}
\end{figure}

Figure~\ref{fig:appm} illustrates the overall workflow of APPM. APPM first identifies an anchor adapter, then performs anchor-protected weighted merging of LoRA matrices, and finally consolidates the frequency gates via softmax competition in the frequency domain.

\subsection{Anchor-Protected Parameter Merging}



A central challenge in adapter merging is that different tasks push shared parameters in opposing directions, so naive averaging produces destructive interference. Rather than treating all task adapters equally, APPM first selects an anchor adapter serving as the reference during merging, since later stages accumulate more knowledge and exhibit stronger representation capability. APPM therefore estimates the importance of each adapter by its aggregated parameter norm:

\begin{equation}
N_k
=
\sum_{l=1}^{L}
\left(
\|
\boldsymbol{U}^{(k)}_l
\|_F^2
+
\|
\boldsymbol{V}^{(k)}_l
\|_F^2
+
\|
\boldsymbol{g}^{(k)}_l
\|_2^2
\right).
\end{equation}

The anchor adapter is selected as

\begin{equation}
\tau^{*}
=
\arg\max_{k}
N_k.
\end{equation}

Once the anchor is identified, APPM aggregates parameters via weighted averaging, where each non-anchor adapter contributes in proportion to its relative parameter magnitude:

\begin{equation}
\lambda_k
=
1
-
\rho
\cdot
\min
\left(
1,
\frac{
\|
\boldsymbol{\Theta}^{(\tau^{*})}
\|_2
}
{
\|
\boldsymbol{\Theta}^{(k)}
\|_2
}
\right),
\end{equation}

where $\rho$ controls the strength of anchor protection.

The merged LoRA parameters are computed as

\begin{equation}
\boldsymbol{\Theta}^{*}_{\text{LoRA}}
=
\frac{
\boldsymbol{\Theta}^{(\tau^{*})}
+
\sum_{k\neq\tau^{*}}
\lambda_k
\boldsymbol{\Theta}^{(k)}
}{
1
+
\sum_{k\neq\tau^{*}}
\lambda_k
}.
\end{equation}

This formulation preserves the dominant knowledge of the anchor while incorporating complementary information from other tasks into the final merged adapter.

\subsection{Frequency-Domain Gate Competition}

Unlike the LoRA matrices, the gate parameters in FA-LoRA determine the
activation of frequency components and therefore directly influence the
frequency characteristics of the learned adapter. Simple parameter averaging
may suppress important task-specific frequency responses.

To preserve informative frequency components, APPM formulates gate merging as a
frequency-wise competition among task adapters. Specifically, the gate
activations are first transformed into probability space through the sigmoid
function. Each task then competes for every frequency component using a
temperature-controlled softmax:

\begin{equation}
\mathbf{p}_k
=
\operatorname{softmax}
\left(
\frac{
\sigma(\boldsymbol{g}^{(k)})
\cdot
\boldsymbol{\beta}_k
}{T}
\right),
\label{eq:gate_softmax}
\end{equation}

where $\boldsymbol{\beta}_k$ denotes the anchor-aware importance coefficient and
$T$ controls the sharpness of the competition.

The merged gate parameters are obtained by

\begin{equation}
\boldsymbol{g}^{*}
=
\sigma^{-1}
\left(
\sum_{k=1}^{K}
\mathbf{p}_k
\odot
\sigma(\boldsymbol{g}^{(k)})
\right),
\label{eq:gate_merge}
\end{equation}

where $\sigma^{-1}(\cdot)$ denotes the inverse sigmoid function.

This competition mechanism assigns each frequency component to the task that
contributes the strongest activation while preserving the dominant role of the
anchor adapter, thereby reducing frequency-domain interference among different
tasks.

\subsection{Computational Complexity}

APPM is performed only once after continual learning has finished and therefore
introduces no additional training cost. The computational complexity is
dominated by parameter aggregation across all adapters:

\begin{equation}
\mathcal{O}
\left(
K
L
(dr+r)
\right),
\end{equation}

where $K$ is the number of continual learning tasks, $L$ is the number of
adapter-augmented transformer layers, $d$ denotes the hidden dimension, and
$r$ is the LoRA rank.

Since APPM operates entirely in parameter space without requiring gradient
computation or training data, it introduces negligible deployment overhead.
Algorithm~\ref{alg:appm} summarizes the complete merging procedure.

\begin{algorithm}[t]
\caption{Anchor-Protected Progressive Merging}
\label{alg:appm}
\DontPrintSemicolon
\SetAlgoLined

Initialize $K$ task adapters $\{\boldsymbol{\Theta}^{(1)}, \ldots, \boldsymbol{\Theta}^{(K)}\}$.

Set protection coefficient $\rho$, anchor coefficient $\beta$, and temperature $T$.

    \textcolor{blue}{\textit{\#\#\# Anchor Identification \#\#\#}}

    \For{each task $k = 1,\dots,K$}
    {
        Compute parameter norm:
        $N_k \leftarrow \sum_l \big( \|\boldsymbol{U}^{(k)}_l\|_F^2 + \|\boldsymbol{V}^{(k)}_l\|_F^2 + \|\boldsymbol{g}^{(k)}_l\|_2^2 \big)$.
    }

    Select anchor: $\tau^{*} \leftarrow \arg\max_k N_k$.

    \textcolor{blue}{\textit{\#\#\# Weighted LoRA Merging \#\#\#}}

    \For{each non-anchor task $k \neq \tau^{*}$}
    {
        Compute contribution weight:
        $\lambda_k \leftarrow 1 - \rho \cdot \min\!\big(1, \|\boldsymbol{\Theta}^{(\tau^{*})}\|_2 / \|\boldsymbol{\Theta}^{(k)}\|_2\big)$.
    }

    Merge LoRA parameters via weighted averaging using $\{\lambda_k\}$.

    \textcolor{blue}{\textit{\#\#\# Frequency Gate Competition \#\#\#}}

    Compute per-frequency softmax allocation $\mathbf{p}_k$ using \eqref{eq:gate_softmax}.

    Merge frequency gates using \eqref{eq:gate_merge}.

    \textcolor{blue}{\textit{\#\#\# Final Assembly \#\#\#}}

    Construct merged adapter:
    $\boldsymbol{\Theta}^{*} \leftarrow \{\boldsymbol{\Theta}^{*}_{\text{LoRA}}, \boldsymbol{g}^{*}\}$.

\Return{$\boldsymbol{\Theta}^{*}$}

\end{algorithm}

\section{Experimental Evaluation}
\label{sec:experiments}

\subsection{Experimental Setup}

We implement all experiments on a server with two NVIDIA RTX A6000 GPUs, each with 48 GB of memory, and one RTX 4090 with 24 GB, using Python 3.10 with PyTorch 2.7.1. The base model is LLaMA-3.2-1B~\cite{grattafiori2024llama} with 4-bit NF4 quantization~\cite{dettmers2023qlora}. Detailed hyperparameter settings are provided in Table~\ref{tab:training_config}.

\textbf{Dataset and Temporal Task Construction.} We use the DIVE benchmark, which contains 22,330 real-world smart contracts with 8 multi-label vulnerability categories~\cite{alsunaidi2026dive}. To construct a deployment-realistic CL evaluation, we sort all contracts by their commit or deployment dates and partition them into $K=4$ contiguous time intervals, yielding tasks task\_A through task\_D, consistent with the temporal constraint of Section~\ref{sec:cl-problem}. As shown in Table~\ref{tab:dataset}, the four tasks have comparable sizes but shifting vulnerability category distributions. Input features are serialized as structured text with a maximum of 1,024 tokens.

\begin{table}[t]
    \renewcommand{\arraystretch}{1.5}
    \caption{DIVE benchmark statistics.}
    \label{tab:dataset}
    \resizebox{\columnwidth}{!}{
    \begin{tabular}{l|rrrr}
    \toprule[1.5pt]
    \rowcolor{gray!10}
    \hline
    \textbf{Split} & \textbf{task\_A} & \textbf{task\_B} & \textbf{task\_C} & \textbf{task\_D} \\
    \hline
        Train & 5,262 & 5,262 & 5,262 & 5,262 \\
        Validation & 542 & 504 & 536 & 651 \\
        Test & 530 & 542 & 513 & 648 \\
        \hline
        Total & 6,334 & 6,308 & 6,311 & 6,561 \\
    \bottomrule[1.5pt]
    \end{tabular}
    }
\end{table}

\textbf{Training.} All CL methods train sequentially on task\_A through task\_D using the AdamW optimizer~\cite{loshchilov2019adamw}. Each task produces a compact FA-LoRA adapter state. The complete hyperparameter configuration is summarized in Table~\ref{tab:training_config}.

\begin{table}[t]
    \renewcommand{\arraystretch}{1.5}
    \caption{Hyperparameter Configuration}
    \label{tab:training_config}
    \resizebox{\columnwidth}{!}{
    \begin{tabular}{m{5.5cm}|>{\centering\arraybackslash}m{2.5cm}}
    \toprule[1.5pt]
    \rowcolor{gray!10}
    \hline
    \multicolumn{1}{c|}{\textbf{Hyperparameter}} & \multicolumn{1}{c}{\textbf{Value}} \\
    \hline
        FA-LoRA rank $r$~\cite{hu2022lora}  & 16  \\
        Frequency mode~\cite{borse2024foura}  & High-frequency retention \\
        Retain fraction $\gamma$~\cite{borse2024foura}  & 0.2 \\
        \hline
        Optimizer~\cite{loshchilov2019adamw}  & AdamW \\
        Learning rate (CL)~\cite{hu2022lora}  & $5 \times 10^{-5}$ \\
        Learning rate (PEFT)~\cite{hu2022lora}  & $3 \times 10^{-5}$ \\
        Batch size $B$~\cite{hu2022lora}  & 8 \\
        Max sequence length  & 1,024 \\
        Epochs per task (CL)  & 3--5 \\
        Epochs (PEFT)  & 3 \\
        Replay buffer capacity~\cite{buzzega2020dark}  & 2,000 \\
        Replay batch ratio~\cite{buzzega2020dark}  & 0.25 \\
        FAR temperature $\tau$  & 2.0 \\
        \hline
        APPM protection $\rho$  & 1.0 \\
    \bottomrule[1.5pt]
    \end{tabular}
    }
\end{table}

\textbf{Evaluation Metrics.} We report per-task and average Micro-F1, FWT and BWT as defined in Section~\ref{sec:cl}, merge time in milliseconds, CPU memory delta and GPU peak inference memory in MB, and parameter count. We additionally report the Independent per-task upper bound from separate per-task FA-LoRA training, and the No-Adapter lower bound of an untrained model.

\subsection{Parameter Efficiency}

Table~\ref{tab:params} summarizes the parameter efficiency of FA-LoRA. With only 5.2M trainable parameters, FA-LoRA achieves strong detection accuracy while requiring approximately 10 MB of storage per task adapter. The frozen base model accounts for 99.6\% of parameters, shared across all tasks.

\begin{table}[t]
    \renewcommand{\arraystretch}{1.5}
    \caption{Parameter breakdown of FA-LoRA.}
    \label{tab:params}
    \resizebox{\columnwidth}{!}{
    \begin{tabular}{l|rr}
    \toprule[1.5pt]
    \rowcolor{gray!10}
    \hline
    \textbf{Component} & \textbf{Params (M)} & \textbf{Fraction} \\
    \hline
        Total & 1,241.0 & 100.0\% \\
        Frozen (LLaMA base) & 1,235.8 & 99.6\% \\
        Trainable (FA-LoRA) & 5.2 & 0.4\% \\
        \quad — LoRA ($\boldsymbol{U}, \boldsymbol{V}$) & 5.0 & 0.40\% \\
        \quad — Gates ($\boldsymbol{g}$) & 0.2 & 0.02\% \\
        \hline
        Per-task storage & $\sim$10 MB & \\
    \bottomrule[1.5pt]
    \end{tabular}
    }
\end{table}

\subsection{Comparison with PEFT Baselines}

Before evaluating continual learning performance, we first validate the effectiveness of FA-LoRA as a parameter-efficient fine-tuning method by comparing it against state-of-the-art PEFT approaches on the full DIVE dataset \textit{without} task splitting. This experiment uses the complete 22,330-contract dataset with all 8 vulnerability labels, evaluating single-task detection capability in isolation from CL concerns. All methods share identical optimization settings as detailed in Table~\ref{tab:training_config}. We evaluate on both LLaMA-3.2-1B and LLaMA-3.2-3B to assess scaling behavior.

We compare seven methods spanning the PEFT design space, including standard LoRA~\cite{hu2022lora}, QLoRA~\cite{dettmers2023qlora}, SLoRA~\cite{huang2026paravul}, FourierFT~\cite{gao2024fourierft}, FouRA~\cite{borse2024foura}, and WaRA~\cite{heidari2025wara}, as well as our proposed FA-LoRA. FourierFT and FouRA are two frequency-domain baselines that share the same architecture and differ only in which frequency band is preserved, providing a controlled ablation that isolates the effect of frequency band selection. WaRA performs low-rank adaptation in the wavelet domain. Detailed configurations are listed in Table~\ref{tab:training_config}.

Table~\ref{tab:peft_comparison} reports Micro-F1, Macro-F1, and subset accuracy for all methods on both model scales, alongside trainable parameter counts and on-disk adapter storage size. FA-LoRA is our proposed method, while FouRA and FourierFT serve as frequency-domain baselines that together form a controlled ablation of frequency band selection.

\begin{table*}[t]
    \renewcommand{\arraystretch}{1.5}
    \caption{Comparison of PEFT methods on the full DIVE dataset without task splitting.}
    \label{tab:peft_comparison}
    \resizebox{\textwidth}{!}{
    \begin{tabular}{l|c|c|c|rr|rr|rr}
    \toprule[1.5pt]
    \multirow{2}{*}{\textbf{Method}} & \multirow{2}{*}{\textbf{Quant.}} & \multirow{2}{*}{\textbf{Trainable (M)}} & \multirow{2}{*}{\textbf{Storage (MB)}} & \multicolumn{2}{|c|}{\cellcolor{gray!10}\textbf{Micro-F1}} & \multicolumn{2}{c|}{\cellcolor{gray!10}\textbf{Macro-F1}} & \multicolumn{2}{c}{\cellcolor{gray!10}\textbf{Subset Acc.}} \\
    \cmidrule(lr){5-6} \cmidrule(lr){7-8} \cmidrule(lr){9-10}
    & & & & \textbf{1B} & \textbf{3B} & \textbf{1B} & \textbf{3B} & \textbf{1B} & \textbf{3B} \\
    \hline
        WaRA~\cite{heidari2025wara} & \checkmark & 35.77 & 136.5 & \textbf{0.8398} & \textbf{0.8515} & \textbf{0.6529} & \textbf{0.7133} & \textbf{0.5544} & \textbf{0.5840} \\
        QLoRA~\cite{dettmers2023qlora} & \checkmark & 1.72 & 6.6 & \underline{0.8185} & 0.8365 & \underline{0.6133} & 0.6407 & \underline{0.5181} & 0.5298 \\
        SLoRA~\cite{huang2026paravul} & \checkmark & 6.85 & 26.2 & 0.8138 & 0.8305 & 0.5945 & 0.6263 & 0.4966 & 0.5262 \\
        LoRA~\cite{hu2022lora} & bf16 & 3.42 & 13.0 & 0.8094 & 0.8370 & 0.5613 & 0.6356 & 0.4845 & 0.5428 \\
        FourierFT~\cite{gao2024fourierft} & \checkmark & 0.16 & 0.6 & 0.7449 & 0.7888 & 0.4724 & 0.5585 & 0.3471 & 0.4331 \\
        FouRA~\cite{borse2024foura} & \checkmark & 0.55 & 2.1 & 0.7635 & 0.8020 & 0.4998 & 0.5930 & 0.4004 & 0.4648 \\
        \hline
        \textbf{FA-LoRA} & \checkmark & 2.62 & 10.0 & \underline{0.8185} & \underline{0.8424} & 0.5904 & \underline{0.6616} & 0.5074 & \underline{0.5544} \\
    \bottomrule[1.5pt]
    \end{tabular}
    }
\end{table*}

Table~\ref{tab:peft_comparison} reports the comparison at both model scales. FA-LoRA attains accuracy close to the strongest baseline while training only 0.4\% of the parameters, and at the 3B scale it outperforms both LoRA and QLoRA across all three metrics using 23\% fewer parameters than LoRA. This advantage stems from retaining the most informative high-frequency components, which concentrates adapter capacity on vulnerability-relevant patterns. The FouRA versus FourierFT comparison confirms the effect, since FouRA preserves high frequencies and outperforms FourierFT at both scales.

\subsection{Comparison of Continual Learning Algorithms}

\begin{figure}[t]
    \centering
    \includegraphics[width=\linewidth]{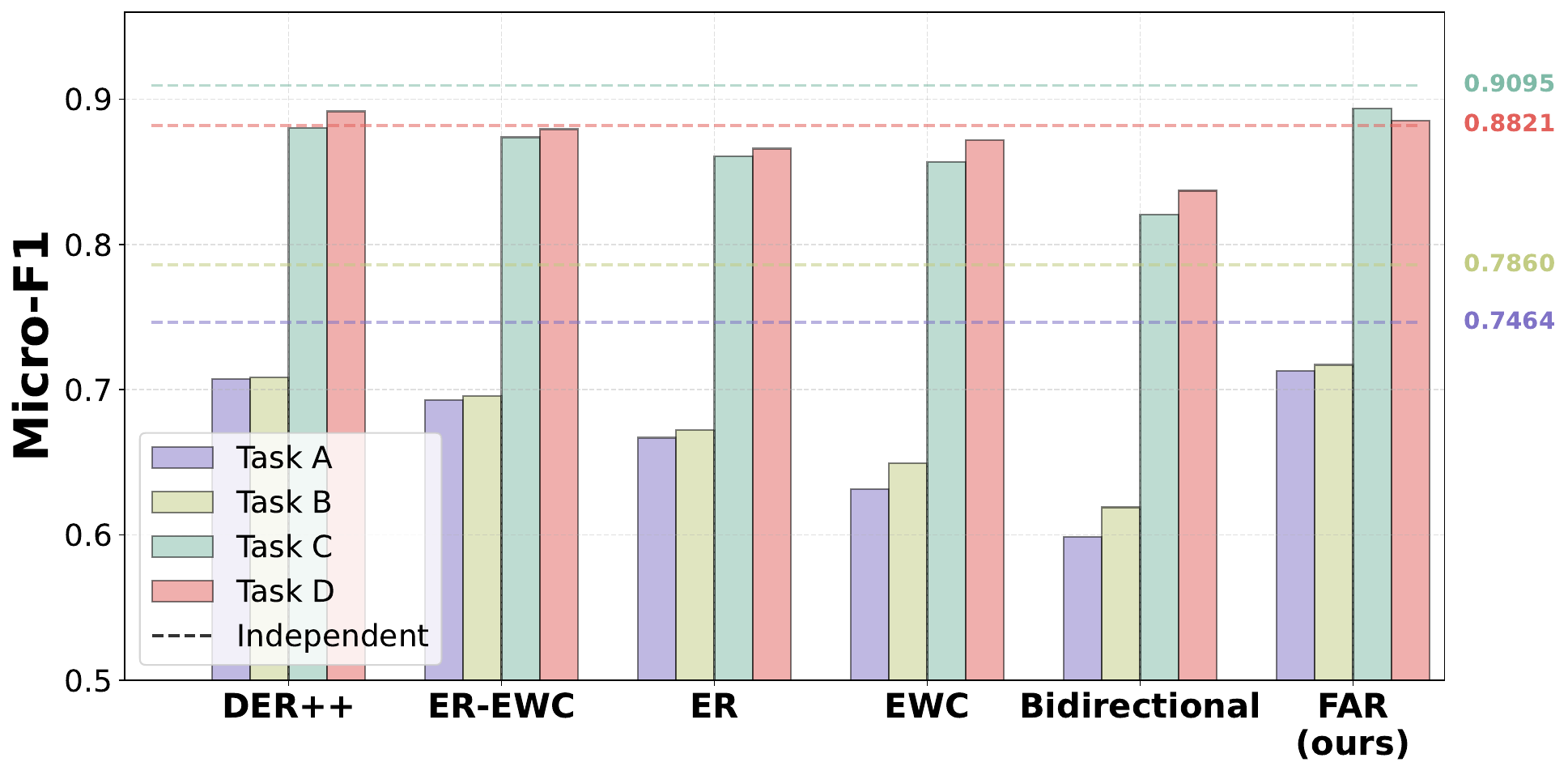}
    \caption{Continual learning algorithm comparison on the DIVE benchmark. Grouped bars show per-task Micro-F1 for each method, with dashed lines indicating the Independent single-task upper bound.}
    \label{fig:cl_comparison}
\end{figure}

Fig.~\ref{fig:cl_comparison} presents the comprehensive comparison of CL algorithms on the DIVE benchmark, reporting per-task Micro-F1 after all four tasks have been learned. The Independent upper bound is shown as dashed reference lines.

Fig.~\ref{fig:cl_comparison} reveals several key findings:

As shown in Fig.~\ref{fig:cl_comparison}, FAR achieves the best overall CL performance with an average Micro-F1 of 0.8022, outperforming all five baselines and trailing the Independent upper bound by only 3.5\%. The loss-dynamics-driven prioritization ensures that the most vulnerable knowledge receives the most rehearsal, yielding stronger retention than uniform replay with identical buffer capacity.

\subsection{Merging Performance}

\begin{figure}[t]
    \centering
    \includegraphics[width=\linewidth]{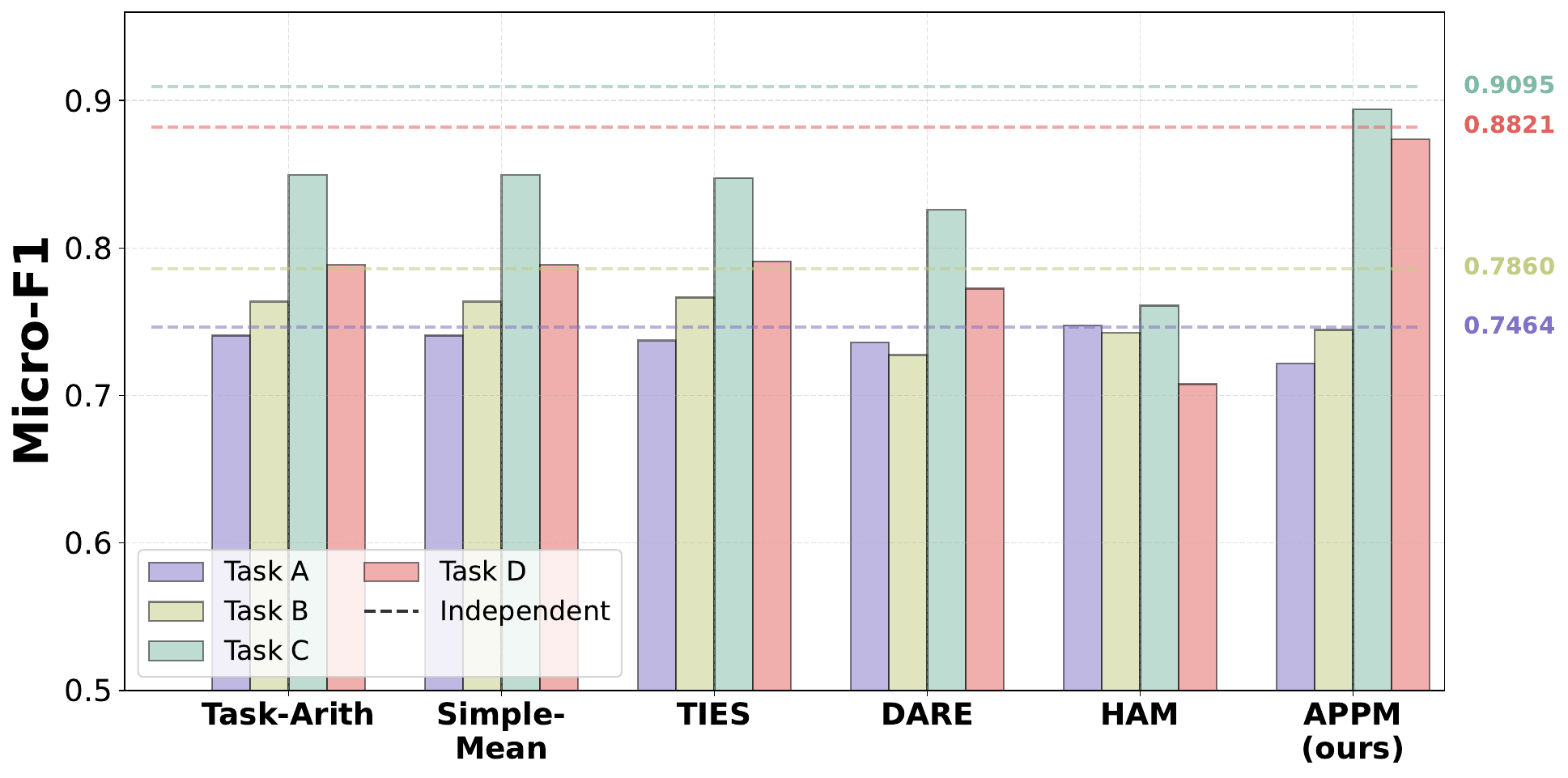}
    \caption{APPM merging comparison on the DIVE benchmark. Grouped bars show per-task Micro-F1 for each merging method, with dashed lines indicating the Independent single-task upper bound. APPM achieves the best average Micro-F1 of 0.8085, closest to the Independent upper bound.}
    \label{fig:merge_comparison}
\end{figure}

Fig.~\ref{fig:merge_comparison} presents APPM against baseline merging methods. All methods merge the four task-specific FA-LoRA adapters into a single adapter without accessing training data. Dashed lines mark the Independent upper bound, while dotted lines mark Sequential. Table~\ref{tab:merge_resources} provides detailed resource metrics.

\begin{table}[ht]
    \renewcommand{\arraystretch}{1.5}
    \caption{Resource metrics for adapter merging methods. All methods produce identical model architectures.}
    \label{tab:merge_resources}
    \resizebox{\columnwidth}{!}{
    \begin{tabular}{l|rrrr}
    \toprule[1.5pt]
    \rowcolor{gray!10}
    \hline
    \textbf{Method} & \textbf{Merge (ms)} & \textbf{CPU $\Delta$(MB)} & \textbf{Speedup} & \textbf{$\Delta$Ind} \\
    \hline
        Simple-Mean           & 102   & 0.0  & 70$\times$  & +5.5\% \\
        TIES~\cite{yadav2023ties}     & 2,123 & 0.0  & 3.4$\times$ & +5.5\% \\
        DARE~\cite{yu2024dare}        & 4,204 & 0.0  & 1.7$\times$ & +7.9\% \\
        HAM g=2                & 7,164 & 68.1 & 1.0$\times$ & +11.0\% \\
        SFA a=0.5              & 72    & 0.0  & 100$\times$ & +33.0\% \\
        \hline
        \textbf{APPM (ours)}   & 156   & 0.0  & 46$\times$  & \textbf{+2.7\%} \\
    \bottomrule[1.5pt]
    \end{tabular}
    }
\end{table}

As shown in Fig.~\ref{fig:merge_comparison} and Table~\ref{tab:merge_resources}, APPM achieves near-optimal multi-task inference with an average Micro-F1 of 0.8085, within 2.7\% of the Independent upper bound. On task\_C and task\_D, APPM outperforms all baseline merging methods, which shows that anchor protection preserves the strong generalization of later adapters. APPM completes merging in 156 ms with zero additional runtime memory overhead.

\subsection{Ablation Studies}

\subsubsection{APPM Component Ablation}

\begin{figure}[t]
    \centering
    \includegraphics[width=\columnwidth]{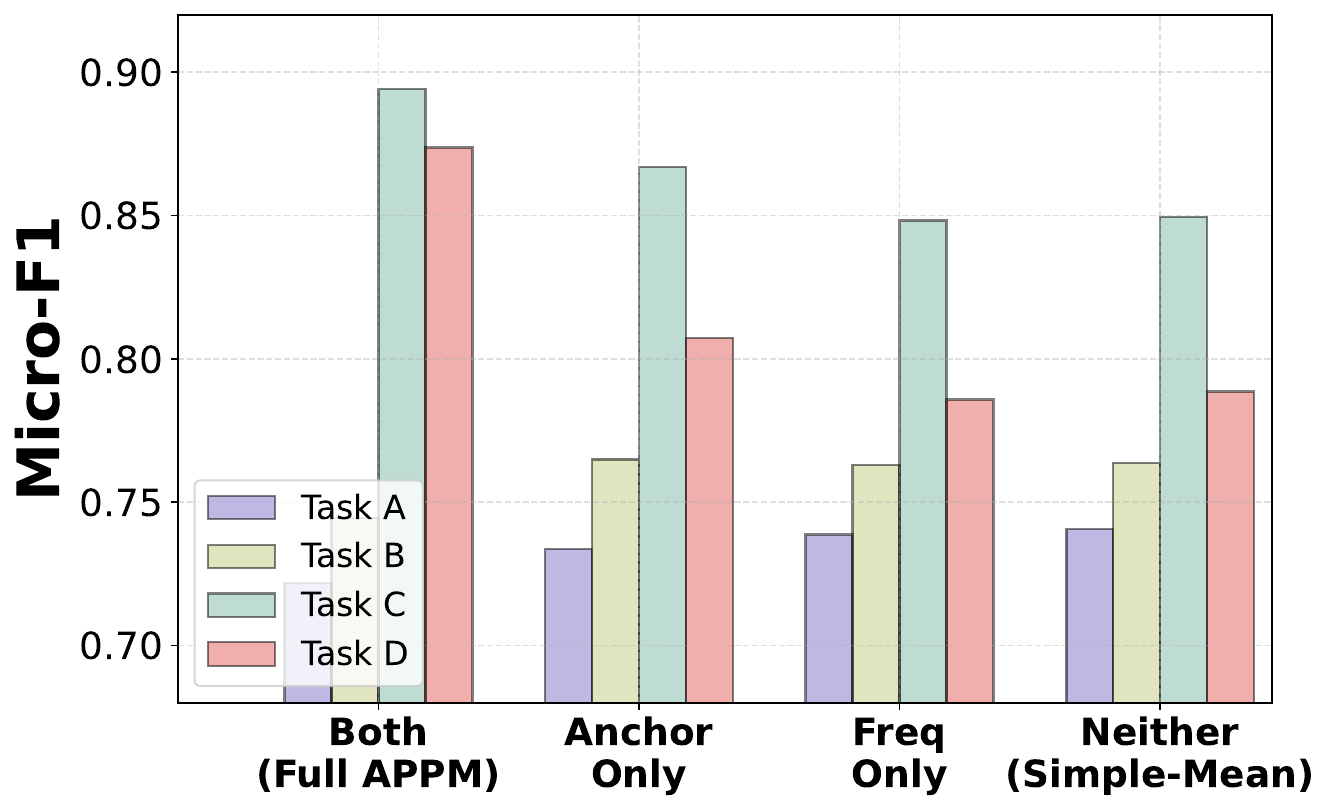}
    \caption{APPM component ablation. All variants merge 4 FAR adapters. The full APPM configuration achieves the best balance across all four tasks, while removing either anchor protection or frequency competition degrades performance, so both mechanisms are necessary.}
    \label{fig:ablation}
\end{figure}

We ablate the two key mechanisms of APPM: anchor protection for LoRA parameters and frequency-domain gate competition. Fig.~\ref{fig:ablation} reports results with all four component combinations. Each variant merges the same four FAR adapters, where Anchor indicates whether task\_D, the identified anchor, is protected and Freq indicates whether frequency-domain softmax competition is used. The Anchor-only variant corresponds to norm-weighted parameter averaging, and the Freq-only variant corresponds to SAM-Orthogonal merging.

\begin{table}[]
    \renewcommand{\arraystretch}{1.5}
    \caption{Backward evaluation matrix for FAR. Bold entries denote in-task F1, entries below the diagonal denote backward transfer, and entries above the diagonal denote forward transfer.}
    \label{tab:backward_eval}
    \resizebox{\columnwidth}{!}{
    \begin{tabular}{l|cccc}
    \toprule[1.5pt]
    \rowcolor{gray!10}
    \hline
    \textbf{After training} & \textbf{task\_A} & \textbf{task\_B} & \textbf{task\_C} & \textbf{task\_D} \\
    \hline
        task\_A & \textbf{0.7495} & 0.6121 & 0.6331 & 0.6028 \\
        task\_B & 0.7253 & \textbf{0.7503} & 0.7303 & 0.7065 \\
        task\_C & 0.7162 & 0.7226 & \textbf{0.8974} & 0.8864 \\
        task\_D & 0.7128 & 0.7171 & 0.8935 & \textbf{0.8854} \\
        \hline
        Forgetting & $-$0.0367 & $-$0.0332 & $-$0.0039 & — \\
    \bottomrule[1.5pt]
    \end{tabular}
    }
\end{table}
\begin{figure*}[]
\centering
\subfigure[Sensitivity to LoRA rank $r$.]{
    \begin{minipage}[]{0.45\linewidth}
    \centering
    \includegraphics[width=0.95\linewidth]{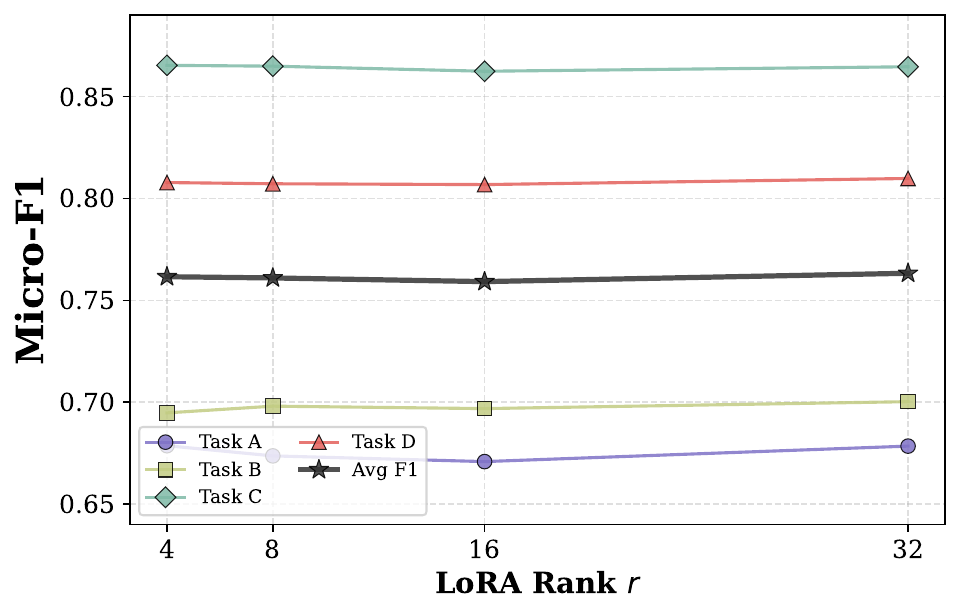}
    \label{fig:sens_rank}
    \end{minipage}
}\hfill
\subfigure[Sensitivity to frequency retention $\gamma$.]{
    \begin{minipage}[]{0.45\linewidth}
    \centering
    \includegraphics[width=0.95\linewidth]{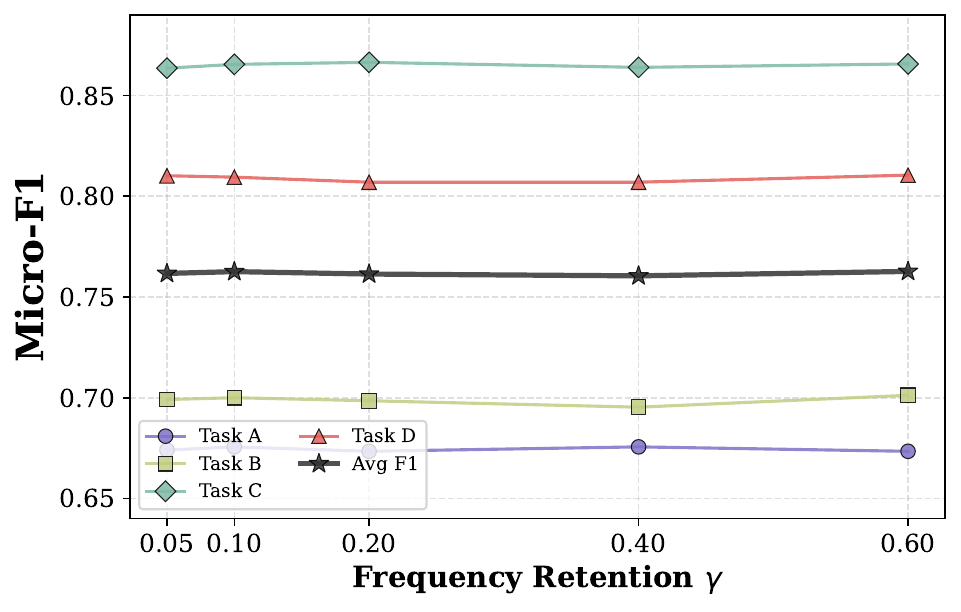}
    \label{fig:sens_gamma}
    \end{minipage}
}\\[0.3cm]
\subfigure[Sensitivity to APPM protection $\rho$.]{
    \begin{minipage}[]{0.45\linewidth}
    \centering
    \includegraphics[width=0.95\linewidth]{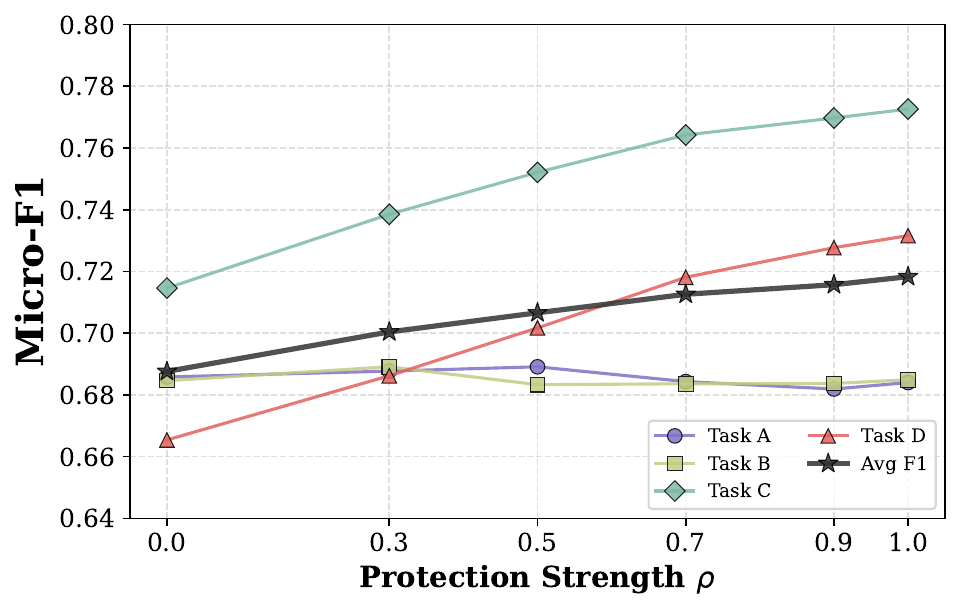}
    \label{fig:sens_rho}
    \end{minipage}
}\hfill
\subfigure[Sensitivity to FAR temperature $\tau$.]{
    \begin{minipage}[]{0.45\linewidth}
    \centering
    \includegraphics[width=0.95\linewidth]{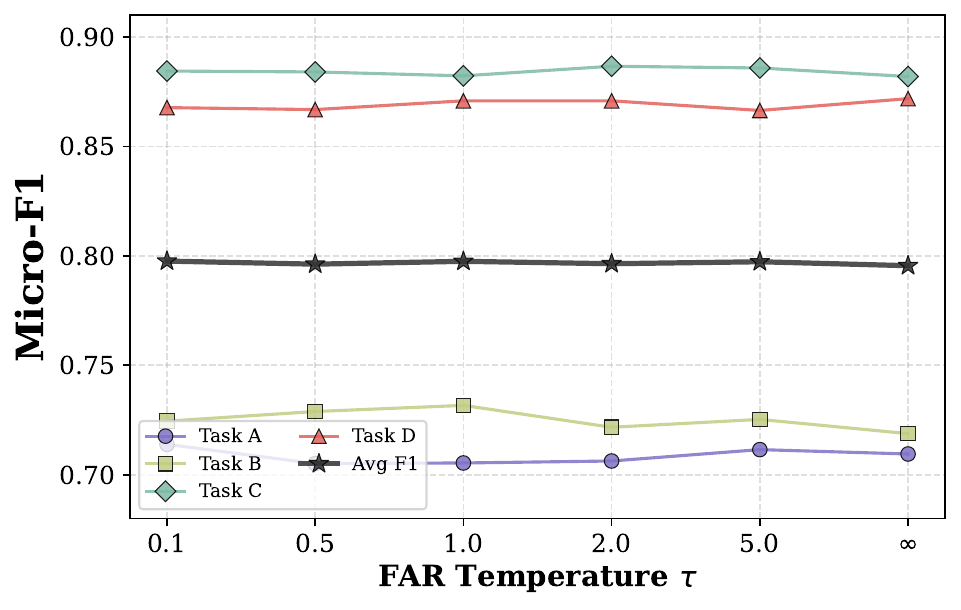}
    \label{fig:sens_tau}
    \end{minipage}
}
\caption{Sensitivity analysis across four hyperparameter dimensions. (a) LoRA rank $r$ is robust with $\Delta$F1${=}$0.0041. (b) Frequency retention $\gamma$ is similarly insensitive with $\Delta$F1${=}$0.0022. (c) APPM protection $\rho$ is the only impactful parameter, yielding ${+}3.07$\% monotonic improvement. (d) FAR temperature $\tau$ has negligible impact with $\Delta$F1${=}$0.0021.}
\label{fig:sens_2x2}
\end{figure*}
\subsection{Sensitivity Analysis}
\label{sec:sensitivity}
Removing anchor protection causes the largest drop on later tasks, where task\_D falls sharply from 0.8737 to 0.8073. Protecting the generalization of the anchor is therefore the primary mechanism. Removing frequency competition instead hurts earlier tasks, where task\_A rises from 0.7218 to 0.7387 as the anchor-free variant over-favors early tasks at the cost of later ones. Removing both mechanisms defaults to Simple-Mean, which occupies a middle ground but sacrifices 2.8\% overall. Only the full APPM configuration successfully balances early-task retention with late-task generalization.

\subsubsection{Backward Evaluation Matrix}

Table~\ref{tab:backward_eval} reports the stage-wise evaluation matrix for FAR. Each row gives the performance on all four tasks after training up to that stage. In-task performance appears on the diagonal, forgetting appears below it, and forward transfer appears above it.


The matrix reveals that forgetting is concentrated in the earliest task. task\_A drops from 0.7495 to 0.7128, a 4.9\% relative decline, whereas task\_C drops by only 0.4\%. This pattern is characteristic of replay-based CL, as the fixed-capacity buffer progressively undersamples earlier tasks when more tasks are added.

We evaluate the sensitivity of four key hyperparameters: FA-LoRA rank $r$, retention ratio $\gamma$, APPM protection strength $\rho$, and FAR temperature $\tau$. A total of 27 experiments are run, each measured by average Micro-F1 across all four tasks. Per-task results are shown in Fig.~\ref{fig:sens_2x2}.

FA-LoRA exhibits strong robustness to rank $r$, with a performance difference of only 0.0041 between $r{=}4$ and $r{=}32$. Frequency-domain gating therefore effectively compensates for reduced rank. Similarly, the frequency retention ratio $\gamma$ shows negligible sensitivity, where even $\gamma{=}0.05$ achieves 0.7617 average Micro-F1, nearly matching $\gamma{=}0.60$ at 0.7627. High-frequency components therefore carry the dominant information. The FAR temperature $\tau$ is also effectively a no-op parameter, as all $\tau$ values span only 0.0021 average Micro-F1.

In contrast, the APPM protection strength $\rho$ is the most impactful hyperparameter. Setting $\rho{=}1.0$ outperforms $\rho{=}0.0$ by ${+}3.07$\% average Micro-F1, with monotonic improvement concentrated on later tasks, where task\_C improves by ${+}5.80$\% and task\_D by ${+}6.62$\%. Overall, we recommend default values for $r$, $\gamma$, and $\tau$, while $\rho$ should be set to its maximum.

\section{Conclusion}
\label{sec:conclusion}

In this paper, we proposed a unified continual learning framework for LLM-based smart contract vulnerability detection that integrates efficient adaptation, forgetting mitigation, and deployment consolidation into a cohesive three-stage pipeline. FA-LoRA achieves competitive detection accuracy with only 0.4\% trainable parameters by retaining high-frequency components in the Fourier domain. FAR tracks per-sample loss dynamics to prioritize vulnerable knowledge for rehearsal, attaining the best CL performance among compared methods. APPM consolidates task-specific adapters via anchor-protected merging with frequency-domain gate competition, achieving performance within 2.7\% of the independent per-task upper bound at millisecond merge cost. Extensive experiments validate the effectiveness of each component and demonstrate that frequency-domain continual learning provides an effective approach for evolving smart contract security. For future work, 6G space-air-ground integrated networks form a hierarchical architecture with heterogeneous computation, communication, and storage capabilities, where edge-cloud collaboration and on-device small language models with LLMs offloaded at edge servers introduce asymmetric resource constraints that may affect the adaptation, replay, and merging stages of our framework. Extending FA-LoRA, FAR, and APPM to such heterogeneous environments is a promising direction for future research.
\bibliographystyle{IEEEtran}
\bibliography{mybibliography}

\end{document}